\documentclass[conference]{IEEEtran}
\IEEEoverridecommandlockouts
\usepackage{cite}
\usepackage{amsmath,amssymb,amsfonts}
\usepackage{graphicx}
\usepackage{textcomp}
\usepackage{xcolor}
\usepackage{mathrsfs}
\usepackage{caption}
\usepackage{graphicx}
\usepackage{float} 
\usepackage{subcaption}
\usepackage{multirow}

\makeatletter
\newif\if@restonecol
\makeatother

\usepackage[linesnumbered,ruled]{algorithm2e}

\def\BibTeX{{\rm B\kern-.05em{\sc i\kern-.025em b}\kern-.08em
    T\kern-.1667em\lower.7ex\hbox{E}\kern-.125emX}}
    
\begin{document}
    
\title{MGIC: A Multi-Label Gradient Inversion Attack based on Canny Edge Detection on Federated Learning}
\author{\IEEEauthorblockN{1\textsuperscript{st} Can Liu}
\IEEEauthorblockA{\textit{Soochow University} \\
\textit{School of Computer and Technology}\\
Soochow, China\\
cliu7@stu.suda.edu.cn}
\and
\IEEEauthorblockN{2\textsuperscript{nd} Jin Wang}
\IEEEauthorblockA{\textit{Soochow University} \\
\textit{School of Future Science and Engineering}\\
Soochow, China\\
wjin1985@suda.edu.cn}
}
\maketitle

\begin{abstract}
As a new distributed computing framework that can protect data privacy, federated learning (FL) has attracted more and more attention in recent years. It receives gradients from users to train the global model and releases the trained global model to working users. Nonetheless, the gradient inversion (GI) attack reflects the risk of privacy leakage in federated learning. Attackers only need to use gradients through hundreds of thousands of simple iterations to obtain relatively accurate private data stored on users' local devices. For this, some works propose simple but effective strategies to obtain user data under a single-label dataset. However, these strategies induce a satisfactory visual effect of the inversion image at the expense of higher time costs. Due to the semantic limitation of a single label, the image obtained by gradient inversion may have semantic errors. We present a novel gradient inversion strategy based on canny edge detection (MGIC) in both the multi-label and single-label datasets. To reduce semantic errors caused by a single label, we add new convolution layers' blocks in the trained model to obtain the image's multi-label. Through multi-label representation, serious semantic errors in inversion images are reduced. Then, we analyze the impact of parameters on the difficulty of input image reconstruction and discuss how image multi-subjects affect the inversion performance. Our proposed strategy has better visual inversion image results than the most widely used ones, saving more than 78\% of time costs in the ImageNet dataset.

\end{abstract}


\section{Introduction}
\label{sec:intro}
To satisfy people's requirements for individual data security, Google~\cite{konevcny2016federated} proposed a federated learning (FL) framework for protecting user privacy in which users only have to upload the trained model's gradients or model weights to the server so that the user's data does not leak into the user's local device. The FL is a distributed computing system framework comprising one server and numerous users~\cite{FedAVG, FedSGD, ovi2023comprehensive, zhu2019deep, zhao2020idlg, geiping2020inverting, yin2021see, dong2021deep,  hatamizadeh2021towards, hatamizadeh2023gradient, zhu2020r, chen2021understanding}. The server distributes a global model to a subset of users the server selected. After users receive the global model, they train it with their local data and upload the model's gradients or model weights to the server. The server updates the global model based on the Federated Averaging (FedAVG)~\cite{FedAVG} or  Federated Stochastic Gradient Descent (FedSGD)~\cite{FedSGD} algorithm with all uploaded gradients or model weights. After that, the server sends the new global model to the selected users in the next round. The whole FL framework repeats the above steps until the global model converges. The server finally obtains the trained global model, the same as the model trained on the user's devices and data~\cite{konevcny2016federated}.  

\begin{figure}[t]
  \centering
   \includegraphics[width=1\linewidth]{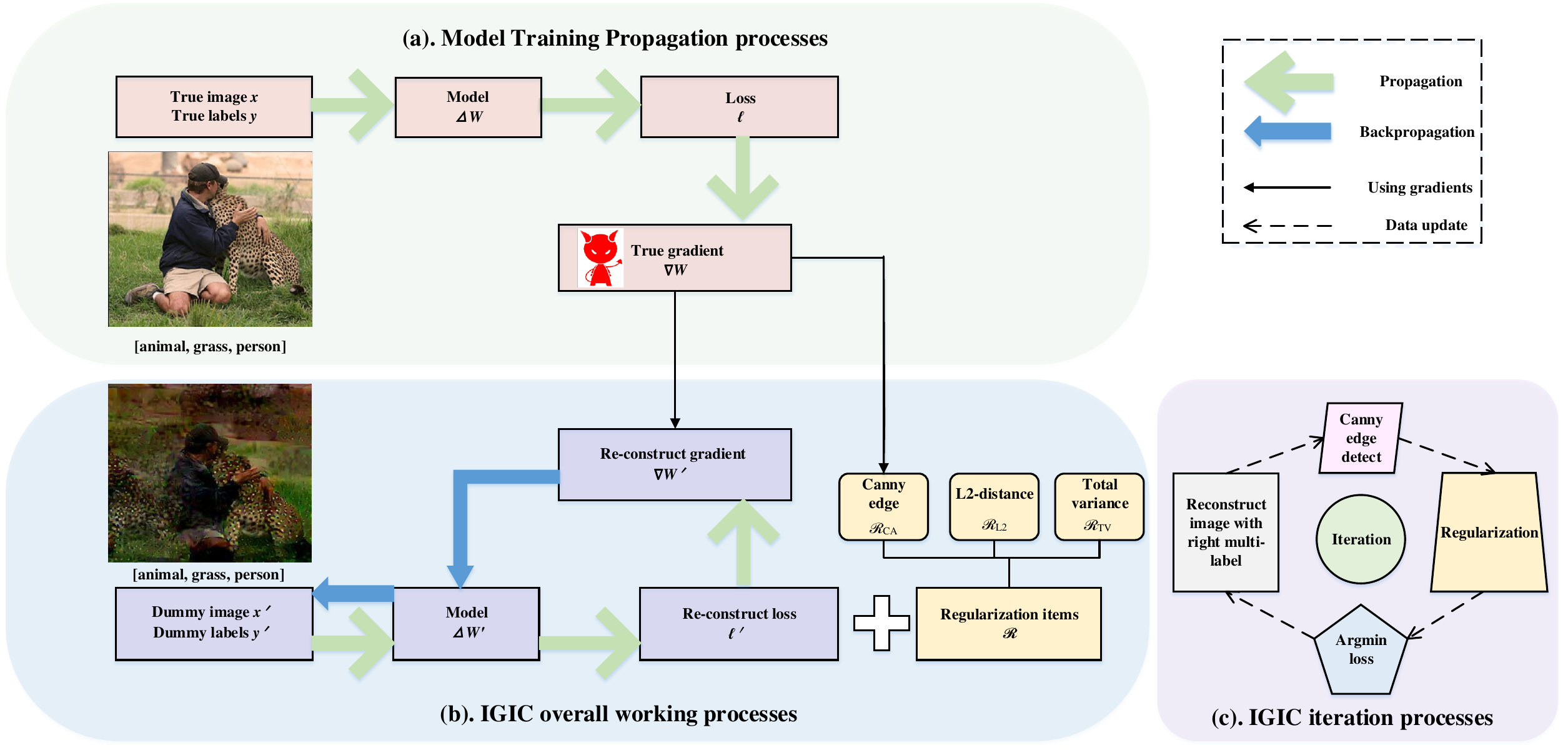}
   \caption{The processes of gradient inversion. In the typical training model, attackers will steal the gradients during the regular training process (green arrows). Attackers could build a similar image with the ground truth image by the model's backpropagation (blue arrows). Attackers will add regularization items (yellow boxes) into the loss function to acccelerate the object function coverage speed and retain the reconstructed image more naturally.}
   \label{fig:inv}
\end{figure}

Although the FL framework is built to protect the privacy of users' local data, recent studies~\cite{zhu2019deep, zhao2020idlg, geiping2020inverting, yin2021see, dong2021deep,hatamizadeh2021towards, hatamizadeh2023gradient, zhu2020r,chen2021understanding} have proved that FL has the risk of privacy leakage. User uploading gradients or model weights to the server has the same effect on the server updating the global model~\cite{konevcny2016federated, lyu2020threats, FedAVG, FedSGD}. Therefore, some researchers~\cite{zhu2019deep, zhao2020idlg, geiping2020inverting, yin2021see, dong2021deep,hatamizadeh2021towards, hatamizadeh2023gradient, zhu2020r,chen2021understanding} assume that gradients as the data passed between the user and the server in the FL framework. The typical FL framework requires users to upload local model gradients to update the global model. There are two kinds of $\mathbf{G}$radient $\mathbf{I}$nversion (GI) attacks are proposed~\cite{zhu2019deep, zhao2020idlg, geiping2020inverting, yin2021see, dong2021deep,hatamizadeh2021towards, hatamizadeh2023gradient, zhu2020r, chen2021understanding}. One is to reduce the distance between the ground truth image and the reconstructed image through multiple iterations~\cite{zhu2019deep, zhao2020idlg, geiping2020inverting, yin2021see, dong2021deep,hatamizadeh2021towards, hatamizadeh2023gradient}. This type of GI attack is an iteration-based attack strategy. Researchers use the users' uploaded gradients to obtain the user's local data through multiple model backpropagations. Geiping et al. (GGI)~\cite{geiping2020inverting}, Yin et al. (GradInversion)~\cite{yin2021see}, Dong et al. (DCI)~\cite{dong2021deep}, and Hatamizadeh et al.~\cite{hatamizadeh2021towards, hatamizadeh2023gradient} add different regularization items to the objective function to obtain more realistic and accurate reconstructed images. This kind of GI attack is simple but effective in obtaining local user data that should be hidden. The other is the recursion-based method~\cite{zhu2020r,chen2021understanding}. It exploits the relationship between the input data, model parameters, and gradients. At present, most researchers are still conducting research on iteration-based GI attacks.

Existing GI attacks are performed on single-label datasets~\cite{zhu2019deep, zhao2020idlg, geiping2020inverting, yin2021see, dong2021deep,  hatamizadeh2021towards, hatamizadeh2023gradient, zhu2020r, chen2021understanding}, such as ImageNet, MINIST, CIFAR-10~\cite{deng2009imagenet, xiao2017fashion, recht2018cifar}, etc. The unique label of an image can be determined by calculating the cross-entropy value. However, the number of images with only one subject is much smaller than that of multiple subjects. In the image classification problem, researchers have introduced the method of multi-label classification to express the image semantics more completely~\cite{bucak2010multi,zhang2013review,chen2019learning,qi2020mlrsnet,wang2021can,yun2021re, gao2021learning}. Researchers propose several multi-label datasets and publish them to detect the accuracy of multi-label classification algorithms, e.g., multi-label high spatial resolution remote sensing dataset (MLRSNet)~\cite{qi2020mlrsnet}, the nus-wide dataset~\cite{chua2009nus}, etc. 
 
Under the image classification task, it is essential to find all subjects in the image~\cite{chen2019learning,yun2021re, gao2021learning}. To find image's all subjects, many edge detection algorithms have been proposed~\cite{ canny1986computational, chen2012laplacian, shrivakshan2012comparison, yan2017method, li2022improved}. The canny algorithm is a standard algorithm for edge detection and is still widely used in computer vision research. This detection algorithm is limited by the direction of the gradients, non-maximum suppression, and two thresholds to describe the edge~\cite{ canny1986computational}. It is considered the best algorithm for edge detection and much more accurate in identifying the edges of images. Therefore, many improved algorithms based on canny edge detection have been proposed in recent years~\cite{xu2017canny, yan2017method, li2022improved}.

This paper proposes a new attack method to obtain high-quality reconstructed images with lower time costs. We assume that users' local data are images so as to make GI attack results more straightforward. In Fig.~\ref{fig:inv} (a) and (b) are the model training propagation and iteration-based GI backpropagation processes, respectively. Minimizing the loss function value between the ground truth and adding regularization in the reconstructed image can close the reconstructed image to the ground truth image. We propose a new iteration-based $\mathbf{M}$ulti-label $\mathbf{GI}$ attack strategy based on the $\mathbf{C}$anny edge detection (MGIC). MGIC aims to achieve lower time costs without losing more or even better reconstructed image quality in single-label and multi-label datasets. According to the canny edge detection research, we can fully use the obtained gradients' value to fetch the ground truth image information. This information helps us narrow the scope of the ground truth image subject position, speeding up the reconstructed image and increasing the accuracy of the reconstructed image. To resolve semantic errors in GI and be closer to the application scene of the actual image, we introduce the multi-label into GI attacks. Using multiple labels explains more image subjects and reduces the possibility of semantic errors due to the lack of subjects in the GI attack. Nevertheless, how to obtain image multi-labels from gradients is an unexplored problem. Using the cross-entropy value to obtain a single label~\cite{zhao2020idlg} is unsuitable for acquiring the image's multi-label. In this paper, we innovatively add a $\mathbf{N}$ew $\mathbf{C}$onvolutional layers $\mathbf{B}$lock (NCB) based on the existing training model. The gradients are used as the input of the new NCB to obtain the label probability value. If probability values exceeding the set threshold are used as the multi-label of the image. We will describe the multi-label acquisition method in section~\ref{PA}. We use the nus-wide multi-label and ImageNet single-label datasets to prove MGIC's better inversion effect. According to existing studies, the GI attack strategy effective in batch size = 1 training setting is still effective in the batch size $>$ 1. At the same time, the GI results quality will decline. In this way, this paper focuses on the GI attack on batch size = 1.

 Our main contributions are as follows:
\begin{itemize}
\item  First, we are the first to focus on the multi-label classification in the GI attack. After the gradient is processed, a new label probability is obtained as input. We use new NCBs to get the relative accuracy results of multi-label on ImageNet and nus-wide datasets. The multi-label helps us reduce the semantic error of similar images and improve the accuracy of GI attacks. 


\item Second, we applied canny edge detection in the GI attack. This design makes full use of the information on gradients and helps us determine the location of subjects in the inversion image. MGIC's results show that much information contains essential features of the image hidden in gradients.
 
\item Third, MGIC achieves better inversion results with only nearly 20\% of the time cost from GGI. The PSNR and SSIM metric values are better than GGI on the ImageNet dataset. The SSIM metric value of MGIC is significantly higher than that of existing strategies, especially on the nus-wide dataset. 
 
\end{itemize}

 \section {Related Works} 
\label{RW}
 %
In this section, we detail the currently accepted GI attack strategies. We give the different objective functions of previous GI studies in subsection~\ref{GI}. In the subsections~\ref{ML}, we explain the multi-label and introduce a realistic multi-label dataset.

 \subsection{Gradient Inversion (GI) Attack}
 \label{GI}
Under the FL framework, the GI attack has been an effective way to get local privacy data. More studies are researching GI attacks in the iteration-based direction. In 2019, Zhu proposed a method called deep leakage from gradient (DLG) to achieve the goal of inversing gradients to the local data $x$ from users~\cite{zhu2019deep}. This strategy results well in MNIST, CIFAR-100, SVHN, and LFW datasets under the ResNet-56 CNN model. The objective function of DLG is to minimize the $L2$ distance between reconstructed data gradients $\nabla W'$ and ground truth gradients $\nabla W$. The object function is:

\begin{equation}
  \begin{split}
    \mathop{argmin}\limits _{\hat{x}, \hat{y}}\sum  \Vert \nabla W'- \nabla W \Vert ^2.
  \end{split}
\end{equation}

DLG requires hundreds of iterations to achieve visually distinguishable reconstructed images. The initialization of the DLG are the random dummy image data $\hat{x}$ and the label $\hat{y}$. This simple initialization easily leads to algorithm convergence failure or label derivation error as $\hat{y} \ne y$. To solve the wrong label in the DLG strategy, iDLG utilizes the negative cross-entropy value of the correct image label to obtain the correct single-label~\cite{zhao2020idlg}.

Geiping's GI attack strategy (GGI)~\cite{geiping2020inverting} adds a total variation regularization term $\mathscr{R}_{TV}$ based on the DLG strategy as an additional denoise operation. This regularization improves reconstructed image quality in the CIFAR-10 and ImageNet datasets. GGI chooses the cosine similarity to the cost function to pay more attention to the decreasing angles. GGI assumes that the correct single-label $y$ has been obtained that $\hat{y} = y$ by the cross-entropy value. In this method, the author set the parameters $restarts = 8$ and $max iteration = 24000$ to obtain the reconstructed image. In this way, GGI has high time costs. The objective function of GGI is finding input $\hat{x}$ to minimize the value of the formula:
\begin{equation}
  \begin{split}
    \mathop{argmin}\limits _{\hat{x}} 1 - cos( \nabla W'- \nabla W ) + \alpha \mathscr{R}_{TV} (\hat{x}),
  \end{split}
\end{equation}
\begin{equation}
\label{cos}
   \cos( \nabla W', \nabla W ) =  \frac{< \nabla W, \nabla W'>}{\Vert  \nabla W\Vert  \Vert  \nabla W' \Vert},
\end{equation}
where the $\cos( \nabla W', \nabla W )$ is the cosine similarity between the reconstructed image gradients $\nabla W'$ and the ground truth gradients $\nabla W$. 

Under the batch size = 1 training model, user privacy leakage (CPL)~\cite{wei2020framework} effectively attacks inversion ground truth images. It implements efficient GI attacks by minimizing the $L2$ distance between the reconstructed image gradients $\nabla W'$ and the ground truth gradients $\nabla W$. CPL adds a label-based regularization $f$ to guarantee the attacking model convergence. The goal function of the CPL is:
\begin{equation}
  \begin{split}
    \mathop{argmin}\limits _{\hat{x}} \Vert \nabla W'- \nabla W \Vert ^2 + \alpha \Vert f( \hat{x}, W' )- y \Vert ^2.
  \end{split}
\end{equation}

Inspired by the inversion effect of  $\mathscr{R}_{TV}$, GradInversion introduces two types of regularization terms, fidelity regularization ($\mathscr{R}_{fidelity}$) and group regularization ($\mathscr{R}_{group}$)~\cite{yin2021see}. $\mathscr{R}_{fidelity}$ includes $\mathscr{R}_{TV}$, $\mathscr{R}_{L2}$ and a new term called the Batch normalization regularization ($\mathscr{R}_{BN}$). $\mathscr{R}_{group}$ helps GradInversion get the reconstructed image object's position. It applies in the FL framework with batch size $>$ 1 and gets better inversion results. The overall goal of GradInversion is:
\begin{equation}
  \begin{split}
    \mathop{argmin}\limits _{\hat{x}} \Vert \nabla W'- \nabla W \Vert ^2 +  \mathscr{R}_{fidelity}(\hat{x}) + \mathscr{R}_{group}(\hat{x}).
  \end{split}
\end{equation}

After GradInversion, the NVIDIA team proposes divide-and-conquer inversion (DCI) with new regularizations based on layers, images, and cycle-consistences regularization that composes the $\mathscr{R}_{img}$ in GANs model~\cite{dong2021deep}. The DCI uses cross-entropy as the cost function between $\hat{x}$ and $\hat{y}$ as $CE( \hat{x}, \hat{y} )$. The goal formula changed into:
\begin{equation}
  \begin{split}
    \mathop{argmin}\limits _{\hat{x}} CE ( \hat{x}, \hat{y} ) +  \sum \mathscr{R}_{BN}(\hat{x}) + \alpha \mathscr{R}_{img}(\hat{x}),
  \end{split}
\end{equation}
where $\mathscr{R}_{BN}(\hat{x})$ corresponds to the BN layers' mean and variance. Then, NVIDIA applies GI with the $\mathscr{R}_{prior}$, $\mathscr{R}_{BN}$ and $\mathscr{R}_{grad}$ in strategy and gets great results in X-ray images~\cite{hatamizadeh2021towards, hatamizadeh2023gradient}.

Although the above strategies obtain visualized reconstructed images in the single-label dataset, they achieve better visual results at higher time costs. They use the cross-entropy value to determine the exact single-label value for a single-label image. However, it can cause severe semantic errors in the reconstructed image during these different GI attack strategies.

Besides the above iteration-based strategies, using recursive-based methods to minimize error in reconstructed data is the other effective GI attack. R-GAP is proposed to reconstruct model layers' input from the last layer~\cite{zhu2020r}. This method obtains the optimal solution with the minimum error by solving several linear formulas. Based on the R-GAP, COPA gives a more profound understanding of the objective function and explains many models' limitations~\cite{chen2021understanding}. Although R-GAP shows that a good visual effect of the reconstructed image can be obtained by iterating 300 times, these two recursive-based methods still focus on image restoration of single-label datasets, which may cause semantic errors in reconstructed images. Therefore, the effectiveness of existing GI attack strategies for realistic multi-agent image reconstruction could be better.

\subsection{Image Multi-label}
\label{ML}

The most popular datasets used in the GI attack are the CIFAR-10 and ImageNet~\cite{zhu2019deep, zhao2020idlg, geiping2020inverting, yin2021see, dong2021deep, hatamizadeh2021towards, hatamizadeh2023gradient}. The common point of the above datasets is that they are all single-label datasets, which means that one image has only one label even though this image has more than one subject. The number of single-subject images is far smaller in real life than in multi-subject images. Thus, more researchers are studying multi-label classification~\cite{bucak2010multi, cheng2010graded, zhang2013review, chen2019learning, yun2021re, gao2021learning}. Chen~\cite{chen2019learning} proposes an SSGRL framework by adding the CNN model structures with a semantic decoupling module to guide learning semantic-specific representations and get the MS-COCO and PASCAL VOC 2007 and 2012 datasets' image multi-label. In~\cite{yun2021re}, Yun proposes a ReLabel strategy to obtain multi-label under the ImageNet dataset. In this way, ImageNet images are classified from single-label to multi-label. ReLabel changes the training model softmax layer and replaces the RoIAlign in the label map. According to the final output probability, researchers could determine the multi-label of the image and the relative positions of label subjects. Gao~\cite{gao2021learning} successfully achieves multi-label annotation of MS-COCO and PASCAL VOC 2007 and 2012 by proposing the MCAR framework with a multi-class attentional region module.

With multi-label getting more studies, multi-label datasets are proposed to apply in the multi-label classification direction~\cite{qi2020mlrsnet, chua2009nus}. The nus-wide dataset, which has 1 to 5 multi-label of one image, is proposed by the National University of Singapore~\cite{chua2009nus} and has been recognized by many relevant researchers. With the advantage of multiple labels' annotation, the semantics of each image is better expressed. The nus-wide dataset has 81 classes and over 250,000 images from real websites. We will use the nus-wide as our simulation multi-label dataset in the following section.

\begin{figure}[t]
  \centering
   \includegraphics[width=1\linewidth]{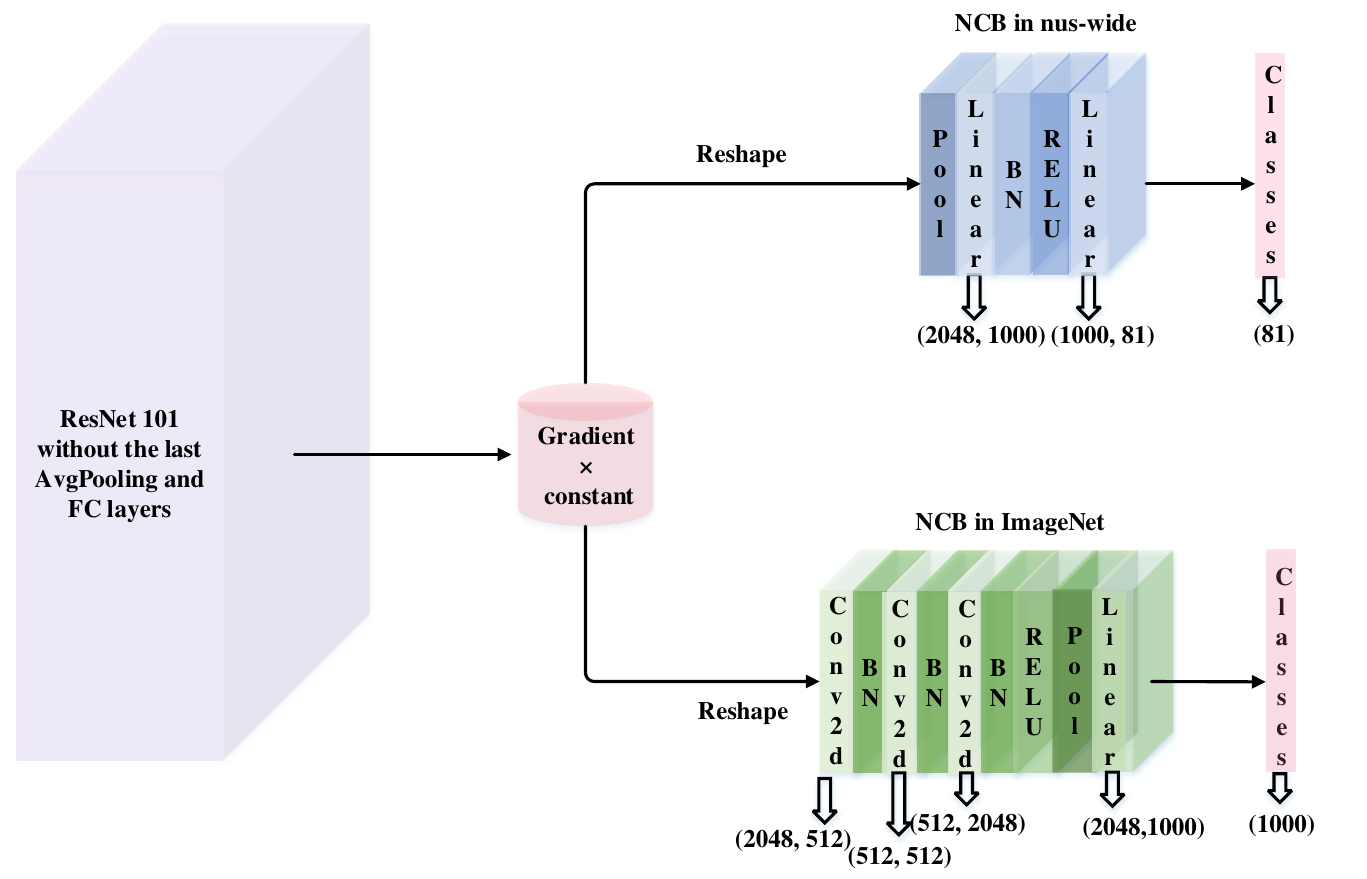}
   \caption{Identify multi-label classification on nus-wide and ImageNet datasets. The ResNet-101 model is changed without the last AvgPooling layer and FC layer. After the image propagation through the new ResNet101 model, the attacker gets the image gradients. The NCB input is gradients multiplied by a large constant. The NCB on the nus-wide (blue blocks) and in ImageNet (green blocks) is composed as the figure shows. The size of these new layers is below layers, respectively. Finally, we get each label probability of the input data. The label's probability larger than the artificially set threshold could be deemed as the input image multi-label.}
   \label{fig:NCN}
\end{figure}

\section{MGIC Strategy} 
\label{PA}

We give MGIC's objective mathematics function expression in subsection~\ref{OF}. Then, the method and process of multi-label acquisition are described in detail in  subsection~\ref{IM}. The multi-label acquisition and calculation processes of the new regularization $\mathscr{R}_{CA}$ will be stated in the last subsection~\ref{AR}.

\subsection{Object Function}
\label{OF}
According to the existing studies in the GI attack~\cite{zhu2019deep, zhao2020idlg, geiping2020inverting, yin2021see, dong2021deep, hatamizadeh2021towards, hatamizadeh2023gradient}, we follow the general assumption that the attack model is an FL system composed of a server $S$ and $N$ users $u_i, i \in \{0, 1, 2, ... , N-1\}$ in this paper. The attacker is usually an honest-but-curious server $S$. The honest-but-curious server will work honestly with the principle of the FL framework. However, it could store the data during model training processes. The $S$ could get a new reconstructed image $\hat{x}$ under a large number of iterative backpropagation operations, which is closed to the ground truth image $x$~\cite{zhu2019deep, zhao2020idlg, geiping2020inverting, yin2021see, dong2021deep, hatamizadeh2021towards, hatamizadeh2023gradient}. Following the study of the GGI~\cite{ geiping2020inverting}, we choose the cosine similarity as the cost function of MGIC to accelerate convergence speed based on angle information. The objective function of MGIC is:
\begin{equation}
\label{obj}
    \mathop{argmin}\limits _{\hat{x}, \hat{y}}1 - cos( \nabla W', \nabla W ) + \mathscr{R}_{reg}.
\end{equation}
The $\mathscr{R}_{reg}$ has explained in the subsection~\ref{AR}. The goal of MGIC is making \eqref{obj} approaching 0. If \eqref{obj} approaches 0, the $\hat{x}$ is near $x$. In this case, we get the reconstructed image similar to the ground truth image through the MGIC strategy. In Fig.~\ref{fig:inv}(b), we show an overall working diagram of the MGIC. 

\subsection{Identify Multi-label}
\label{IM}
In \eqref{obj}, there are 2 factors $\hat{x}$ and $\hat{y}$ influence the other items $\nabla W', \nabla W$ and $\mathscr{R}_{reg}$. The $\hat{x}$ is the final result we need to resolve, and we cannot determine the range of  the $\hat{x}$ before iterations end. However, the value of $\hat{y}$ can be obtained by gradients. Under the ground truth label $y$, the accuracy and convergence speed could be highly improved~\cite{zhao2020idlg,geiping2020inverting, yin2021see, dong2021deep,hatamizadeh2021towards,  hatamizadeh2023gradient,  zhu2020r,chen2021understanding}. In~\cite{zhao2020idlg}, the negative cross-entropy value has been used to confirm the unique single-label of the image. This method is useful in the single-label ImageNet, CIFAR-10, and CIFAR-100 datasets. Nevertheless, it is not suitable for the acquisition of getting the multi-label. Thus, we need to confirm the image's multi-label first.

Inspired by~\cite{yun2021re, gao2021learning, chua2009nus, chen2019learning} classify the image into the multi-label with a new model added to the training model. We keep the backbone of the ResNet-101 network and only change the output of the final FC layer from 1000 to 81. The nus-wide classification trained model holds the ResNet-101 model before the last AvgPooling (POOL) and FC layers~\cite{chua2009nus}. In~\cite{chua2009nus}, authors add a $\mathbf{N}$ew $\mathbf{C}$onvolutional layers $\mathbf{B}$lock (NCB), which is composed by the new Linear, AvgPooling, Batch-Norm, and FC layers. The detailed structure is shown in Fig.~\ref{fig:NCN} blue blocks to get the multi-label of the nus-wide dataset's image. The weights of NCB have been proved in~\cite{chua2009nus}. However, the input of this NCB in ~\cite{chua2009nus} is the output value after normal network propagation in the remaining ResNet-101 model. As an attacker, $S$ cannot get this output value and only knows gradients. The gradients are derivative of the model output value and the layers' weights. It is easy to find that if the output value is large, the gradients' value of this output should be large as the weights of the model's one layer are constants. To decrease the different iterations' model weights effect and keep the value relationship between gradients' value unchanged, we let the gradients before the last AvgPooling layer in the normal ResNet-101 model multiple a large constant as the new input of the NCB. For example, if $G_A > G_B$ in the gradients matrix and we will get $G_A \times 7 \times 10^8 > G_B \times 7\times 10^8$. The new input of the NCB is $G_A \times 7\times 10^8$ and $G_B \times 7\times 10^8$. Finally, we use the new input to obtain the probability values corresponding to each label. Through the artificial threshold setting, the image's multi-label is the corresponding label whose probability value exceeds the set threshold. The exact value of this threshold will be given in section~\ref{SIM}, and the pseudocode for obtaining the multi-label in the nus-wide is in Algorithm~\ref{IMP}.

Under the ImageNet dataset, we construct a similar NCB as in the nus-wide dataset. The new training model remains the ResNet-101 model before the last AvgPooling layer and adds an NCB, which is composed of the last block of the RestNet-101 model's last layer block (represented as the second sequence of the fourth layer sequence in the ResNet-101 model). The weights of this NCB are the same as the pre-trian ResNet-101 model's corresponding layers. The detailed architecture of ImageNet NCB is shown in~\ref{fig:NCN} green blocks. Through the artificial threshold setting, the image's multi-label is the corresponding label whose probability value exceeds the set threshold. The exact value of the threshold will be given in section~\ref{SIM}. It is worth noting that the ImageNet dataset is single-label. However, we do not set the number of image labels to be one. To better fit the characteristics of the ImageNet images and speed up code execution, we set the maximum number of labels for the ImageNet images to 2. The selected multi-label that exceeds the threshold must correlate significantly with the image.

\begin{algorithm}[t]
  \caption{Identify multi-label in nus-wide}
  \label{IMP}
    \KwIn
      {Before the last POOL layer's Gradients $BPG_n$; NCB in nus-wide $NCB_n$; Threshold $Thre_n$.}
    \KwOut
      {Reconstruct Multi-label $RM_n$.}
     $G'_n \leftarrow$ $BPG_n  \times 7 \times 10^8$;\\
     $G_n \leftarrow G'_n~reshapes~to~[1000, 2048, 1, 1]$;\\
     multi-label's~probability~$MP_n \leftarrow NCB_n(G_n)$;\\   \tcp{$G_n~trains~in~NCB_n$}
    \For {$i~from~0~to~80$}
    {
     	{\If {$MP_n[i] >Thre_n$}
         {$RM_n$ $\leftarrow$ multi-label;}
	}
	}
\end{algorithm}

\begin{algorithm}[h]
  \caption{Canny regularization $\mathscr{R}_{CA}$}
  \label{CA_reg}
    \KwIn
      {Before the last POOl layer's Gradients $BPG$; The elements number of matrix $num(\cdot)$; Max iteration $MI$; Coordinate index $index(\cdot)$; the $i$-th iteration reconstruct image $\hat x_i$; Canny edge detection $CA( \hat x_i, thre_1, thre_2)$; Scaling the matrix's pixel $scale(\cdot)$.}
    \KwOut
      {Canny edge regularization $\mathscr{R}_{CA}$.}
    $fin = (max(BPG)-min(BPG)) \times 0.6$\tcp*{get~the~gradients~threshold}
    \For{$i~from~0~to~num(BPG)$}
    {
    	\If{$BPG[i] > fin$}
    	{$PCG'' \leftarrow index(BPG[i]);$\\ \tcp{select~gradients~coordinates}}
    }
    $PCG' \leftarrow index(PCG''(\lfloor1/2 \times num(PCG'')\rfloor, \lfloor2/3 \times num(PCG'')\rfloor))$;\\
    $PCG \leftarrow scale(PCG')$\tcp*{get~the~coordinate~in~the~image}
	\For{$i~from~0~to~MI$}
	{$P'_{rec}\leftarrow CA(\hat x_i, thre1, thre2);$\\
		$P_{rec}\leftarrow index(P'_{rec}( \lfloor1/2 \times num(P'_{rec}) \rfloor, \lfloor2/3 \times num(P'_{rec}) \rfloor))$\tcp*{ get~coordinates~in~$\hat x_i$}
	    	  $\mathscr{R}_{CA} \leftarrow \Vert P_{rec} - PCG\Vert^2;$
     }
\end{algorithm}

\subsection{Regularization Terms}
\label{AR}

Although the GI attack can inverse images, the GI attack might have a bad result as it takes too long time and the model cannot converge. To speed up model convergence and improve model convergence probability, researchers add different regularization terms to their objective functions~\cite{geiping2020inverting, yin2021see, dong2021deep,  hatamizadeh2021towards, hatamizadeh2023gradient, zhu2020r, chen2021understanding}.

Following these studies, we add regularization terms to get better GI results. Our objective function has shown in~\eqref{obj} and the $\mathscr{R}_{reg}$ is composed of three parts:
 \begin{equation}
\label{reg}
    \\  \mathscr{R}_{reg} = \alpha_{TV}\mathscr{R}_{TV} + \alpha_{L2}\mathscr{R}_{L2} + \alpha_{CA}\mathscr{R}_{CA}.
\end{equation}
\begin{equation}
\label{CA}
    \\  \mathscr{R}_{CA} = \Vert  CA_g -  CA_t\Vert ^2.
\end{equation}
In~\eqref{CA}, $CA_g$ is the pixel's coordinate we select from gradients, and $CA_t$ is the pixel's coordinate obtained from canny edge detection in every GI attack iteration. This subsection's fifth paragraph gives the methods for obtaining these two nodes' coordinates. Use $\mathscr{R}_{CA}$ penalty to reconstruct images that deviate significantly from their position and continuously approach the correct subject's position. The $\alpha_{TV}$, $\alpha_{L2}$ and $\alpha_{CA}$ are the scaling parameters of regularizations, respectively.

The $\mathscr{R}_{TV}$ as an image prior regularization keeps the reconstructed image realistic in both~\cite{geiping2020inverting} and~\cite{yin2021see} GI attack strategies. In~\cite{yin2021see}, $\mathscr{R}_{L2}$ as a penalization of the $L2$ norm of $\hat{x}$. Thus, we still apply $\mathscr{R}_{TV}$ and $\mathscr{R}_{L2}$in MGIC. We do not introduce $\mathscr{R}_{BN}$ as it is time-consuming and impractical in the FL framework~\cite{xu2022agic}.

In~\cite{yin2021see}, researchers have found that the reconstructed image subject's position may deviate from the ground truth. To solve this problem, they introduce $\mathscr{R}_{group}$ into their objective function to take the average of the subject's position from different initial seeds, and $\mathscr{R}_{group}$ alleviates the significant deviation. $\mathscr{R}_{group}$ achieves better reconstructed image quality through more random seeds' initialization. Therefore, using $\mathscr{R}_{group}$ increases the time costs of the GI attack.

To assure the image subjects' positions, we set a new regularization called Canny regularization ($\mathscr{R}_{CA}$) based on the canny edge detection algorithm.
\begin{itemize}
\item The processes to get the  $CA_g$ are:
\begin{itemize}
\item First, we make a threshold $fin$ in dynamic. The value of $fin$ is related to the maximum and minimum values of gradients ($max(gra)$ and $min(gra)$):
\begin{equation}
\label{fin}
    \\  fin = (max(gra)-min(gra)) \times 0.6.
\end{equation}

\item We take the center position of the pixels corresponding to the pixels' gradient values larger than $fin$. We select the 1/2-th and 2/3-th of the selected site pixels' coordinates as the baseline. 

\item However, the gradients' matrix size is not the same as the image, such as the gradients' matrix is a $gra_{row} \times gra_{col}$ matrix and the image data is an $img_{row} \times img_{col}$ matrix. We use proportional scaling to scale the coordinates in the selected gradients matrix to the image matrix. For example, if  we select the $[i, j]~(i \in \{0, ..., gra_{row}-1\},~j \in \{0, ..., gra_{col}-1\})$ pixel as the baseline point in the gradients matrix, this pixel corresponds to the $[\lfloor i \times img_{row} \setminus gra_{row}\rfloor, \lfloor j \times img_{col} \setminus gra_{col}\rfloor]$ in the image. 
\end{itemize}

\item The  $CA_t$ is getting by using canny edge detection on the reconstructed images in every backpropagation iteration. We choose the middle position's pixel coordinate of the selected edges as the reconstructed baseline point. 

\item Calculate the $L2$ distance of the two points' coordinates and penalize the reconstructed image with a significant difference in the distance between these two points.

\end{itemize}
The application of canny edge detection has been represented in Fig.~\ref{fig:inv}(c). In whole iteration processes, we will use canny edge detection to get the position of the reconstructed image. Then, this detecting will be applied in the $\mathscr{R}_{CA}$ to assure subjects' sites. The specific $\mathscr{R}_{CA}$ pseudocode is in Algorithm~\ref{CA_reg}.

\section{Experiments}
\label{SIM}
We first give the implementation details in our model and strategy in the subsection~\ref{ID}. In subsection~\ref{MR}, we show the success of MGIC and compare it with GGI~\cite{geiping2020inverting}. 

\begin{table}
\renewcommand\arraystretch{1.5}
\tabcolsep=.6cm
	\caption{Metric Comparision in ImageNet}
	\centering
	\label{tab:vali}
	\begin{tabular}{cccc}
		\hline
		\multicolumn{2}{c}{\multirow{2}*{Metrics}} & \multicolumn{2}{c}{Strategies}\\
		\cline{3-4}
		\multicolumn{2}{c}{\multirow{2}*{~}} &{GGI}&{MGIC}\\
		\hline
		\multicolumn{2}{c}{\multirow{2}*{PSNR $\uparrow$}} &{8.14}&{$\mathbf{11.58}$}\\
		& &{7.48}&{$\mathbf{9.88}$}\\
		\hline
		\multicolumn{2}{c}{\multirow{2}*{SSIM $\uparrow$}}&{$\mathbf{2.6005e-2}$}&{2.30e-2}\\
		& &{$\mathbf{6.0268e-2}$}&{3.04e-2}\\
		\hline
		\multicolumn{2}{c}{\multirow{2}*{Objective Function $\downarrow$}}&{0.2132}&{$\mathbf{0.1411}$}\\
		& &{0.1958}&{$\mathbf{0.1461}$}\\
		\hline
	\end{tabular}
\end{table}

\begin{figure}[t]
	\centering
	\begin{subfigure}{0.32\linewidth}
		\centering
		\includegraphics[width=0.9\linewidth]{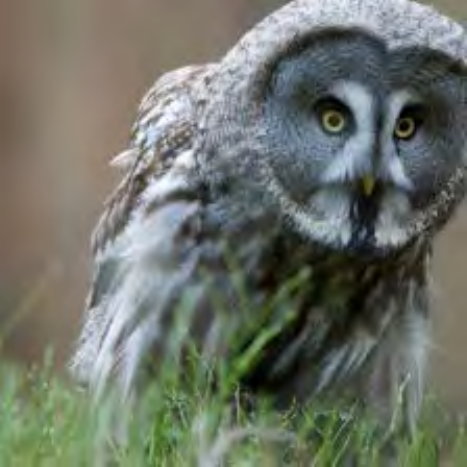}
		\caption{Ground Truth}
		\label{4-a}
	\end{subfigure}
	\centering
	\begin{subfigure}{0.32\linewidth}
		\centering
		\includegraphics[width=0.9\linewidth]{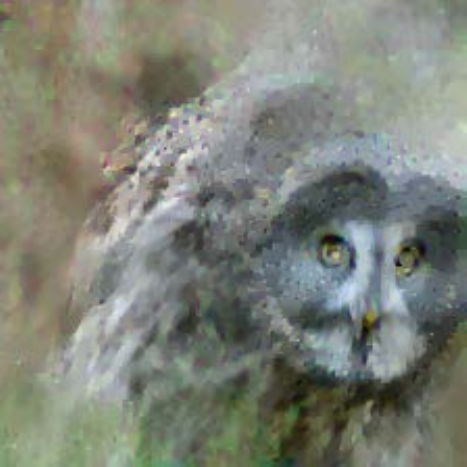}
		\caption{GGI}
		\label{4-c}
	\end{subfigure}
	\centering
	\begin{subfigure}{0.32\linewidth}
		\centering
		\includegraphics[width=0.9\linewidth]{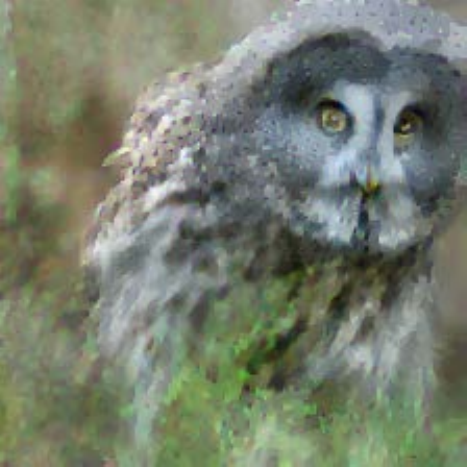}
		\caption{MGIC}
		\label{4-d}
	\end{subfigure}\\
	
	\begin{subfigure}{0.32\linewidth}
		\centering
		\includegraphics[width=0.9\linewidth]{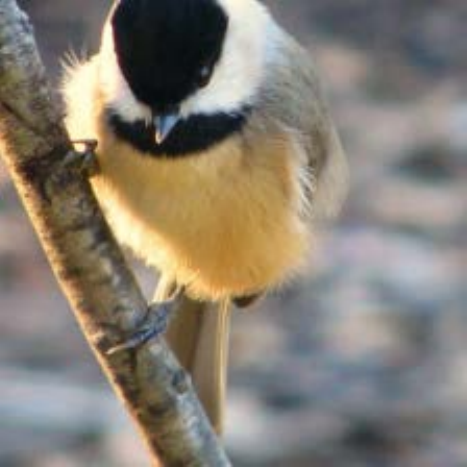}
		\caption{Ground Truth}
		\label{4-e}
	\end{subfigure}
	\centering
	\begin{subfigure}{0.32\linewidth}
		\centering
		\includegraphics[width=0.9\linewidth]{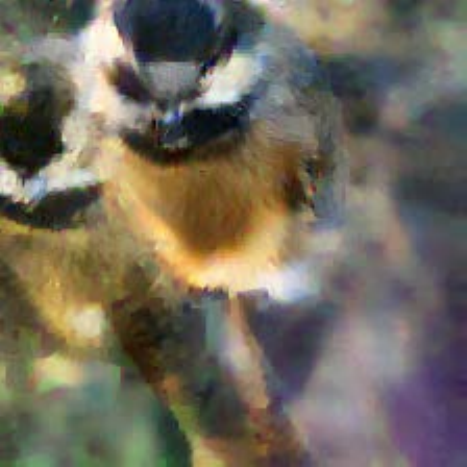}
		\caption{GGI}
		\label{4-f}
	\end{subfigure}
	\centering
	\begin{subfigure}{0.32\linewidth}
		\centering
		\includegraphics[width=0.9\linewidth]{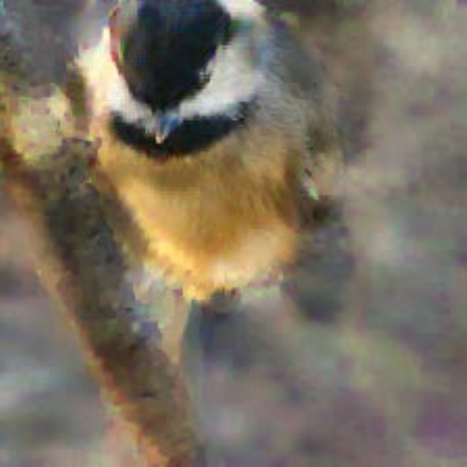}
		\caption{MGIC}
		\label{4-g}
	\end{subfigure}\\
	\caption{Reconstructed images in the ImageNet dataset. Both GGI and MGIC are in the seeting of $restart = 1$ and $max~iteration = 20K$. The leftmost column is the ground truth image.}
	\label{fig:imgi}
\end{figure}

\subsection{Implementation Details}
\label{ID}
In all cases, image pixels are initialized from the Gaussian distribution with the mean = 0 and variance = 1 following the same set of GGI and GradInversion~\cite{geiping2020inverting, yin2021see}. As found in~\cite{geiping2020inverting, yin2021see}, shallower network (ResNet-18 and ResNet-32) inversion results are worse than deeper network (ResNet-101 and ResNet-152). Higher-resolution images and pre-trained models are more challenging to reconstruct. We primarily focus on a shallower pre-trained ResNet-101 architecture and batch size = 1 for the classification task and reconstruct the ground truth image of resolution 224px $\times$ 224px. We use Adam for optimization with a 0.01 learning rate.

After the newly added NCB, we get a multi-label that must contain the right label in ImageNet, under which we set the maximum label number as 2. We get the ground truth label $\hat y$ by the gradient cross-entropy negative value. The other multi-label is the corresponding label with the largest probability value, and the probability value is greater than $0.99 \times \hat y$ in the new NCB. It is worth mentioning that most ImageNet images are still single-label images, as the label's probability is not larger than the set threshold. According to~\cite{chua2009nus}'s research, the number of images containing three types of tags, clouds, person, and sky, exceeds 60\% of the total images. Therefore, we set the maximum label number as 3 in the nus-wide dataset. The NCB label restoration accurately is 100\% in ImageNet and more than 60\% in the nus-wide, which is higher than 34\% through the cross-entropy method. The multi-label obtains at least three ground-truth labels based on the above setting. One of the $\hat y$ is obtained by the cross-entropy value. The other multi-label is the corresponding label with the largest probability value, and the probability value is greater than $0.99 \times \hat y$ in the new NCB. We use $\alpha_{TV} = 1 \times 10^{-1}$ and $\alpha_{L2} = 1 \times 10^{-5}$ which is the same setting as~\cite{geiping2020inverting, yin2021see}. The $\alpha_{CA} = 1 \times 10^{-6}$ both in ImageNet and nus-wide datasets. The threshold of the canny edge detection is related to the reconstructed image. In $\mathscr{R}_{CA}$, the $CA(\hat x, thre1, thre2)$ paremeters are $thre1 = max(\hat x) \times 0.8$ and $thre2 = max(\hat x) \times 0.9$ in dynamic where the $max(\hat x)$ is the maximum value of $\hat x$. We are using the NVIDIA A100 GPU to simulate GGI and MGIC.
 
In GGI~\cite{geiping2020inverting}, the GI attack is setting as $restart = 8$ and $max~iteration = 24K$. Thus, there are $192K$ times iterations in one reconstructed image, and the total time costs are about 8.5 GPU hours. In MGIC, we simulate the GI attack in $restart =1$ and $max~iteration = 20K$. The total iteration in MGIC at most is $40K$ and $60K$ in ImageNet and nus-wide dataset. In the number of iterations, GGI is larger than 4.8 times and 3.2 times that of the MGIC. For a fair comparison of the effects of GGI and MGIC, we set the $max~iteration = 20K$ values of GGI to be the same as those of MGIC. The $restart$ in GGI is the same as the number of each image's multi-label in ImageNet and nus-wide datasets. To present visual comparisons of reconstructed images with GGI, we compute two metrics to measure the image similarity: $(1)$ post-registration PSNR and $(2)$ structural similarity SSIM between ground truth and reconstructed images. The larger the PSNR and SSIM values, the more similar the two pictures are. MGIC uses fewer additional regularization terms and fewer iteration times than~\cite{yin2021see, dong2021deep, hatamizadeh2021towards, hatamizadeh2023gradient}. Since~\cite{yin2021see, dong2021deep, hatamizadeh2021towards, hatamizadeh2023gradient} do not publish the source code and the max-iterations, we are not performing an accurate experiment time comparison.

 \subsection{MGIC Results}
\label{MR}
\begin{figure}[t]
	\centering
	\begin{subfigure}{0.32\linewidth}
		\centering
		\includegraphics[width=0.9\linewidth]{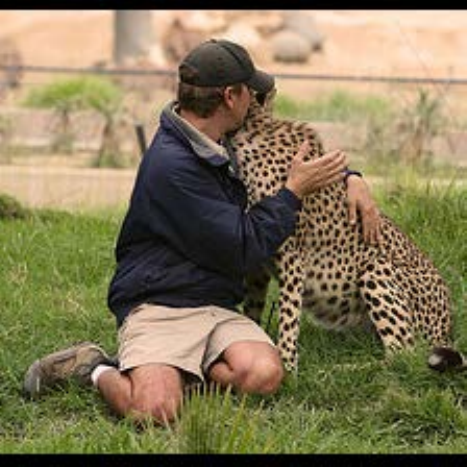}
		\caption{Ground Truth}
		\label{5-a}
	\end{subfigure}
	\centering
	\begin{subfigure}{0.32\linewidth}
		\centering
		\includegraphics[width=0.9\linewidth]{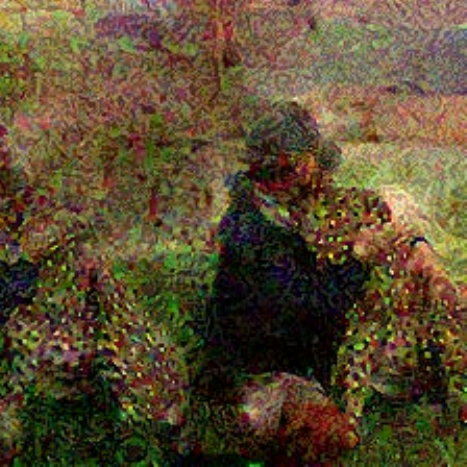}
		\caption{GGI}
		\label{5-c}
	\end{subfigure}
	\centering
	\begin{subfigure}{0.32\linewidth}
		\centering
		\includegraphics[width=0.9\linewidth]{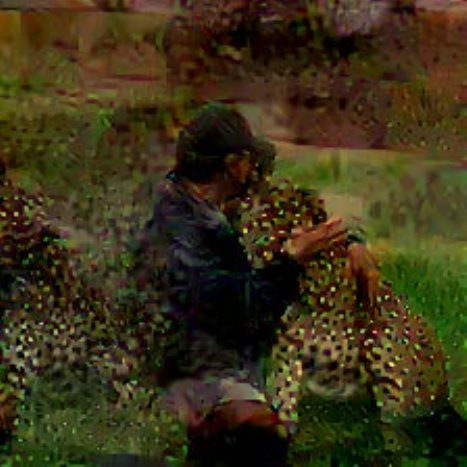}
		\caption{MGIC}
		\label{5-d}
	\end{subfigure}\\
	
	\begin{subfigure}{0.32\linewidth}
		\centering
		\includegraphics[width=0.9\linewidth]{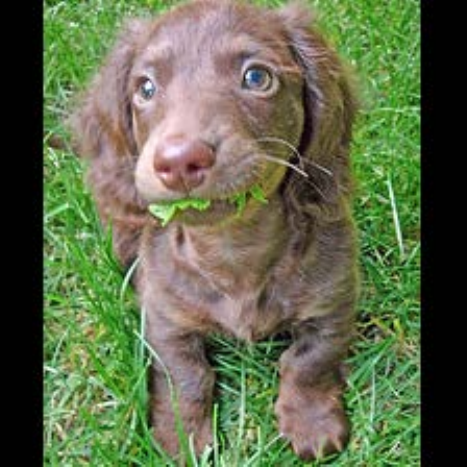}
		\caption{Ground Truth}
		\label{5-e}
	\end{subfigure}
	\centering
	\begin{subfigure}{0.32\linewidth}
		\centering
		\includegraphics[width=0.9\linewidth]{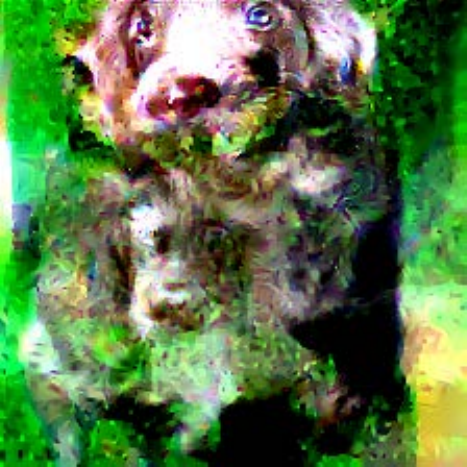}
		\caption{GGI}
		\label{5-f}
	\end{subfigure}
	\centering
	\begin{subfigure}{0.32\linewidth}
		\centering
		\includegraphics[width=0.9\linewidth]{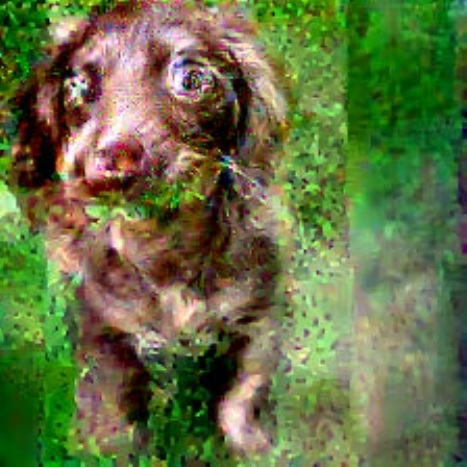}
		\caption{MGIC}
		\label{5-g}
	\end{subfigure}
	\caption{Reconstructed images in the nus-wide dataset. Both GGI and MGIC are in the seeting of $restart = 1$ and $max~iteration = 20K$. The leftmost column is the ground truth image.}
	\label{fig:imgn}
\end{figure}

We summarize the GGI and MGIC results of the ImageNet dataset in the TABLE~\ref{tab:vali}. The corresponding images' visual comparisons are in Fig.~\ref{fig:imgi}. TABLE~\ref{tab:valn} and Fig.~\ref{fig:imgn} are the metrics values and images of the nus-wide dataset. In TABLE~\ref{tab:vali} and TABLE~\ref{tab:valn}, $\uparrow$ means a higher value is better, and $\downarrow$ means a lower value is better.

If the ImageNet gets only one ground truth label through the NCB, MGIC needs 1.38 GPU hours to complete the GI attack. If there are two labels, the GI attack time will be approximately twice that of a single label. These time costs are lower than 8.5 GPU hours. In the ImageNet dataset, PSNR values are higher in MGIC than in GGI. Regarding images' visual effects, the reconstructed images of MGIC are still better than those of GGI. The GGI result in the ImageNet dataset is shown in Fig.~\ref{fig:imgi}(b). The position of the subject of the reconstructed image is shifted in this reconstructed image. In Fig.~\ref{fig:imgi}(e), there are two same subjects in the GGI reconstructed image. This phenomenon is significantly alleviated in MGIC, which is shown in Fig.~\ref{fig:imgi}(c) and Fig.~\ref{fig:imgi}(f). Getting the correct position of the subject is precisely the purpose of introducing the $\mathscr{R}_{CA}$. MGIC is significantly better than GGI in obtaining subject positions and slowing down subject repetition issues.
\begin{table}
\renewcommand\arraystretch{1.5}
\tabcolsep=.6cm
	\caption{Metric Comparision in Nus-wide}
	\centering
	\label{tab:valn}
	\begin{tabular}{cccc}
		\hline
		\multicolumn{2}{c}{\multirow{2}*{Metrics}} & \multicolumn{2}{c}{Strategies}\\
		\cline{3-4}
		\multicolumn{2}{c}{\multirow{2}*{~}} &{GGI}&{MGIC}\\
		\hline
		\multicolumn{2}{c}{\multirow{2}*{PSNR $\uparrow$}} &{$\mathbf{16.16}$}&{15.59}\\
		& &{13.95}&{$\mathbf{16.74}$}\\
		\hline
		\multicolumn{2}{c}{\multirow{2}*{SSIM $\uparrow$}}&{3.7037e-2}&{$\mathbf{4.9127e-2}$}\\
		& &{3.2618e-2}&{$\mathbf{4.1572e-2}$}\\
		\hline
		\multicolumn{2}{c}{\multirow{2}*{Objective Function $\downarrow$}}&{$\mathbf{0.1554}$}&{0.2007}\\
		& &{0.4132}&{$\mathbf{0.1602}$}\\
		\hline
	\end{tabular}
\end{table}

As there are three labels of each nus-wide image, MGIC costs are about 2.8 GPU hours to reconstruct one image and are lower than 8.5 GPU hours. In TABLE~\ref{tab:valn}, it is easy to find that the lower objective function values do not always have better visually reconstructed images. For example, Fig.~\ref{fig:imgn}(b) has a lower objective function value than Fig.~\ref{fig:imgn}(c), but the SSIM value of Fig.~\ref{fig:imgn}(b) is lower than Fig.~\ref{fig:imgn}(c). The SSIM metric value of Fig.~\ref{fig:imgn}(f) is nearly ten times that of Fig.~\ref{fig:imgn}(e), especially. In Fig.~\ref{fig:imgn}(e), there is more than one subject in the GGI reconstructed image. Multi-label may increase the loss function value, but it has a better effect in improving the images' visual effects. Therefore, the multi-label hides image information much more than a single-label. Thus, multi-label can reduce semantic error in reconstructed images. MGIC also reduces the error of subject repetition in reconstructed images.

\section{Conclusion}
We propose MGIC to GI attack to get the multi-label and relative accuracy ground truth image in batch size =1. Compared to GGI, MGIC needs only 20\% of the time to get a better quality reconstructed image. Our paper offers insights that gradients contain the critical information of the ground truth image. It is meaningful that the FL framework is not a safe distributed learning framework and needs to be continuously improved by developers.

In future work, we aim to apply MGIC within batch size $>$1 by simply fine-tuning in many more models. We will also investigate protection strategies against numerous GI attacks in the FL framework.

\bibliographystyle{IEEEtran}
\bibliography{refe}

\end{document}